\documentclass[letterpaper]{article}
\usepackage{aaai18}
\usepackage{times}
\usepackage{helvet}
\usepackage{courier}
\usepackage{url}
\usepackage{graphicx}
\title{Ethical Questions in NLP Research: The (Mis)-Use of Forensic Linguistics}
\author{Akhilesh Sudhakar\textsuperscript{\textdagger}, \ and Anil Kumar Singh\textsuperscript{\textasteriskcentered}\\
\textsuperscript{\textdagger}akhileshs.s4@gmail.com\\
\textsuperscript{\textasteriskcentered}Indian Institute of Technology (BHU), India\\
nlprnd@gmail.com\\
}

\begin{document}

\maketitle
\begin{abstract}
Ideas from forensic linguistics are now being used frequently in Natural Language Processing (NLP), using machine learning techniques. While the role of forensic linguistics was more benign earlier, it is now being used for purposes which are questionable. Certain methods from forensic linguistics are employed, without considering their scientific limitations and ethical concerns. While we take the specific case of forensic linguistics as an example of such trends in NLP and machine learning, the issue is a larger one and present in many other scientific and data-driven domains. We suggest that such trends indicate that some of the applied sciences are exceeding their legal and scientific briefs. We highlight how carelessly implemented practices are serving to short-circuit the due processes of law as well breach ethical codes. 
\end{abstract} 

\section{Introduction}

Quantitative, statistical and machine learning techniques have greatly changed the way applied sciences -- and even core sciences~\cite{Whiteson:2009} -- are being practiced. The extraordinary success of these techniques for many applications has naturally led to their being used more and more for various kinds of purposes. Many of these are quite benign and do not cause any harm or violate ethical principles. However, there are some for which this is not true. At this point in history, we seriously need to rethink the role of these techniques for many purposes.

While this problem may be relatively absent for the case of core sciences (due to their very nature), we argue that they pose a great challenge for the practitioners of applied sciences. In this paper, we study the case of forensic linguistics as an example to demonstrate why we face a great challenge that may radically change the world we live in. The Rule of Law and other principles of legality and ethics, which form the foundation of the modern world (for the collective good), may be at serous risk, due to certain harmful scientific practices. When we propose or take up an NLP or machine learning technique for deploying in real-world situations, it is vital that we understand and appreciate the ramifications it poses. 

We mainly focus on legal aspects of NLP research, taking forensic linguistics as a specific example in this paper, but we also briefly mention scientific limitations. We start with considering the `due  process'\footnote{\url{https://www.law.cornell.edu/anncon/html/amdt5bfrag1\_user.html}} of law as practiced in most countries (section~\ref{due-process}). We then discuss the roles of forensic science in general  (section~\ref{forensic-science}) and forensic linguistics in particular (section~\ref{forensic-linguistics}). We then point out that the careless practice of NLP and machine learning techniques, especially when they deal with real-world data and can affect people, are short-circuiting the due process (section~\ref{short-circuiting}) of law. Thus, they far exceed the legal brief (section~\ref{legal-brief}). Finally, we point out some scientific limitations of the areas on which forensic linguistics relies (section~\ref{scientific-limitations}).

\section{The Due Process}
\label{due-process}

According to Wikipedia, which can be treated today as a resource for commonly held knowledge (with some well-known issues), the Due process is described as follows:

\begin{quote}
Due process~\cite{James-et-al:2014} is the legal requirement that the state must respect all legal rights that are owed to a person. Due process balances the power of law of the land and protects the individual person from it. When a government takes action against a person without following the exact course of the law, this constitutes a Due process violation, which offends the rule of law.
\end{quote}

For a more authoritative source, we can refer to the Fifth and Fourteenth Amendments to the United States Constitution, which the Supreme Court of the United States interprets as providing four protections: procedural Due process (in civil and criminal proceedings), substantive Due process, a prohibition against vague laws, and as the vehicle for the incorporation of the Bill of Rights.

The Legal Information Institute of the Cornell University Law School\footnote{\url{https://www.law.cornell.edu/wex/due\_process}} lists the Due process rights, as far as procedural Due process is concerned. There are many variations of the Due process in different countries, but the major steps involved in the Due process (for criminal cases) can be summarized as follows:

\begin{itemize}
	\item A complaint is made to the legal authorities (e.g. the police)
	\item A case is registered
	\item The case is investigated first by the police
	\item Suspects are identified and, in the presence of circumstantial evidence, a trial is started
	\item Arguments are made and witnesses are called
	\item The quality of the evidence is evaluated, often by expert witnesses
	\item The jury may deliberate (in some countries)
	\item The case is decided by the judge
	\item A punishment is awarded, if conviction happens
	\item The punishment is carried out by a different legal agency
\end{itemize}

These are fundamental parts of the process which every individual is familiar with, but are still being ignored very frequently, especially when using NLP and machine learning for tasks like crime detection and prediction. There are different agencies that have been given the mandate of different steps of the legal process. The suspect is informed, not only of the fact of him/her being a suspect and of the crime having been committed, but also of his/her rights as per the law. Not only is there a prosecutor, but also a defense attorney, who is central to the legitimacy of the Due process. 

Forensic studies, specifically, play a role in a number of the above mentioned stages in the Due process. This role is elaborated in the following section.

\section{The Role of Forensic Science}
\label{forensic-science}

One definition of forensic science is that it is the application of scientific knowledge and methodology to legal problems and criminal investigations\footnote{\url{http://legal-dictionary.thefreedictionary.com/Forensic+Science}}. It is the investigation, explanation and evaluation of legal relevance~\cite{Fraser:2010}. The most important and frequent role of forensic science, however, is in the evaluation of evidence presented as part of the Due process. Two main roles of forensic scientists are:

\begin{itemize}
\item In collection, preservation and analysis of scientific evidence as part of an investigation
\item As expert witnesses during the trial, working either for the prosecution or the defence
\end{itemize}

However, what concern us the most is that some of the current practices using forensic sciences exceed the mandate of forensic sciences itself. The legally mandated role of forensic science is in fact, far more limited than what current practices would suggest. Forensic studies are meant to work within and inside of the Due process, where the suspects are informed of the crime as well as the rights accorded to them upon being suspects (such as the right to remain silent and the right not to self-incriminate\footnote{\url{http://legal-dictionary.thefreedictionary.com/Self-Incrimination}}). The permitted role of forensic sciences does not extend to taking action against the suspect, and certainly not before being proven guilty through the Due process (we address 'precrime', an extreme breach of legal boundaries of forensic sciences, in ensuing sections). With advancement in data-driven analyses and methods, forensic studies today, are heavily relying on quantitative, statistical and machine learning techniques~\cite{Matsushima:2013}. We highlight the case of forensic lingustics in the following section.

\section{The Role of Forensic Linguistics}
\label{forensic-linguistics}

Forensic linguistics is often described as a branch of Applied Linguistics~\cite{Coulthard:2004} or a branch of Sociolinguistics~\cite{Eades:2010,Coulthard:2017}, and is also related to Corpus Linguistics. Eades describes the role of sociolinguistics in the legal process as that of expert witnesses, through legal education, and through investigating the role of language in the perpetuation of inequality in and through the legal process.

\cite{Olsson:2013a} define forensic linguistics as:

\begin{quote}
Any forensic linguistic inquiry or investigation can draw upon any branch of theoretical or applied linguistics in order to analyse the language of some area of human life which has relevance to the law, whether criminal or civil.
\end{quote}

Yet another common definition is:

\begin{quote}
Forensic linguistics is the analysis of language that relates to the law, either as evidence or as legal discourse.
\end{quote}

There is definitely a valid case to be made for forensic linguistics, but the apprehension we put forth is that recently, it has started overstepping its mandated role, as we discuss later. Forensic linguistics is also mentioned as the process of {\bf solving} word crimes~\cite{Olsson:2013b}. Even within its valid role, it can often be misused. The goal of forensic linguistics, as of forensic science in general, is to ``ferret out the innocent from the guilty'. However, in practice, recent happenings show otherwise too. It can, in fact, be used to `create language crimes'~\cite{Shuy:2005}. A quotation from a review of this book by Robert A. Leonard\footnote{\url{https://muse.jhu.edu/article/256783}} is instructive:

\begin{quote}
`[t]he temptation to cut corners and get an ``obviously'' guilty person convicted is sometimes difficult to overcome' ... and leads to creating the `illusion of a crime', hence the title reference to `creating' language crimes, that is, crimes `accomplished through language alone'.
\end{quote}

A related issue is that of ``adversarial interpreting'~\cite{Kredens:2015}, which shows limitations of linguistic evidence. For instance, the existence of different linguistic varieties of many communities which have difficulty in communicating in other varieties (for instance, the standard variety) can also be used against those communities, as described for Aboriginal communities~\cite{Nothen:2007}. The same could be true of other `profiled' communities. Profiling of a criminal has been a long standing practice~\cite{Campbell:2004}, but the use of NLP and machine learning methods can lead to profiling of not just individual suspects (unaware of being suspected of committing the crime and also not told of their rights), but to the profiling of entire communities as well. While on the one hand statistics can be used for studying racial profiling, for example, on the other it can lead to the profiling itself, if enough care is not taken.

\section{Exceeding the Legal Brief}
\label{legal-brief}

The international forensic liguistics community at large, has started showing concern about the ethical and legal ramifications of the field. The Biennial Conference of the International Association of Forensic Linguists\footnote{\url{http://iafl.org/}} is one of the major conferences for forensic linguistics. The 13th version of this conference is on the theme of `New Challenges for Forensic Linguists'\footnote{\url{http://web.letras.up.pt/IAFLPorto2017}}. Its call for papers lists language minorities and the legal system, linguistic disadvantage before the law, courtroom interpreting and translation and Human Rights matters, apart from language policy and language rights. All of these relate to the use of forensic linguistics for justice. There is also a code of ethics for this area\footnote{\url{https://lsaethics.wordpress.com/category/forensic-linguistics-ethics-statement/}}. Despite such a code of ethics being in existence, there are alarming cases of forensic linguistics being used for purposes such as deception detection\footnote{\url{http://www.montclair.edu/chss/linguistics/deception-detection/}} (some kinds of which can be harmless, but others not), counter-terrorism (not for crime investigation but to predict crime) as well as intelligence and surveillance\footnote{\url{http://www.forensiclinguistics.net/POST-PN-0509.pdf}}. This leads to ``the reduction of protective forms of law in the new `risk law.'''~\cite{Kunz:2013}. 

With the advent of multi-modal systems, computer vision and forensics linguistics are now increasingly being linked. The ICMR workshop on Multimedia Forensics and Security\footnote{\url{http://mklab.iti.gr/mfsec2017/call-for-papers/}} has taken cognizance of the prevailing issue and has issued calls for papers on issues which might be problematic from the point of view of justice.

\subsection{Short-circuiting the Due Process}
\label{short-circuiting}

A consequence of the `risk law' (and what we might call `exception law') mentioned in the previous section is that the crime investigation role of forensic scientist (and of forensic linguist) leads to direct action in the form of various kinds of black lists or even no-fly lists\footnote{\url{https://www.aclu.org/know-your-rights/what-do-if-you-think-youre-no-fly-list}}. In many cases (as in the case of no-fly lists), this happens in secret and the suspect, who is not even aware of being a suspect (or of the supposed crime), is subjected to action without the Due process. This results in short-circuiting of the Due process. This means that the roles of the complainant, investigating agency, prosecution, jury, judge and executioner are all eliminated and replaced by the verdicts of a computational system based on statistical operations on raw data. In such an extra-legal process, there is no role for the defense attorney, who is central to the legal system and without whom there can be no legitimacy of the legal system.

\subsection{Enemy Penology when Everyone is a Suspect}

Sussanne Krassman describes the concept of `enemy penology'~\cite{Krassman:2007} as developed by G{\"u}nther Jakobs, a German professor in criminal law. According to this new paradigm:

\begin{quote}
(N)otorious delinquents, since they are incorrigible, have forfeited their status as citizens: for example, habitual criminals like sexual offenders, professionals involved in the so-called organized crime scene or political criminals including today's predominant concern, that is internationally operating terrorists.
\end{quote}

In the presence of tendencies like racial profiling etc. and the deployment of automatic techniques (often in secrecy), this becomes a threat to the existence of a society based on the Rule of Law as commonly understood. Moreover, when applied on big data and with large scale surveillance, it acquires truly dystopian dimensions, when everyone is a suspect.

%\subsection{Creating Language Crimes}

\subsection{Precrime}

The concept of precrime~\cite{Mantello:2016,McCulloc:2015} is not a new idea, as it originates in science fiction before the advent of modern computers. Quoting Haggerty and Ericson~\cite{Haggerty:2000}:

\begin{quote}
(A) world where social control was no longer exclusively authored by human eyes or security forces but relied on a vast electronic ecosystem of sensors and software.
\end{quote}

In the world of precrime, the final obstacle in the practice of the Rule of Law is demolished: that of time. Crime investigation, and forensics, happens before the crime is committed, instead of after it.

The declared norms of, say, forensic linguistics do not argue for precrime, but it is still becoming a practice with various kinds of detection and crime prediction techniques.

\section{Scientific Limitations}
\label{scientific-limitations}

Added to the dangers to the Rule of Law mentioned above, we finally cannot ignore the limitations of empirical science. Most predictions are not accurate enough to be deployed in situations where they can be used to harm people's livelihood, social standing and even lives. Can preventing plausible harm become a justification for possibly harming people? This is a matter that needs much more detailed analysis. While we do not dwell on it at length in this work, we do want to point out that in current times, when technologies are being deployed on a global scale, even an error of one percentage point is too huge to be brushed aside. As we know, we are struggling to find solutions to problems which are very simple for human beings, such as shallow discourse parsing~\cite{Xue:2016} and to understand the meaning of very simple sentences stories\footnote{\url{http://www.coli.uni-saarland.de/~mroth/LSDSem/}}.

%\subsection{Limitations of Linguistics}

%\subsection{Limitations of Quantitative Methods}

%\subsection{Limitations of Machine Learning and NLP}

\section{Conclusion}

We discussed the role of forensic science in general and forensic linguistics in particular in the context of the Rule of Law and the Due process. These roles are intended to be pretty limited, but with the increasing use of machine learning and NLP techniques in addition to older qualitative and quantitative methods, these roles are being carelessly expanded. This is happening at such a scale that, as we argue, they are now far exceeding their legal and scientific briefs. This is already causing harm to many unsuspecting people and if this continues, we should not be surprised by alarming developments, as we have been witnessing recently. Another dangerous trend is that we are entering an age of precrime that goes against the very fabric of civilized society as we understand it, or at least did till a decade or two ago. The role of forensic linguistics should be restricted to studying and evaluating evidence (as part of Due process) and to provide expert witnesses. Further, this should happen only when a crime is committed, and not to predict the future as in the case of precrime. Finally, we cannot ignore the fact that we know too little so far about the human brain, human psychology and human languages themselves. We are still struggling to solve some very simple problems in NLP, for example. We plan to study all these issues in more detail in the future.

\bibliography{aaai.bib}
\bibliographystyle{aaai}

\end{document}